\documentclass[conference]{IEEEtran}
\IEEEoverridecommandlockouts
\usepackage{cite}
\usepackage{amsmath,amssymb,amsfonts}
\usepackage{algorithmic}
\usepackage{graphicx}
\usepackage{textcomp}
\usepackage{xcolor}
\usepackage{booktabs}
\usepackage{float}

\def\BibTeX{{\rm B\kern-.05em{\sc i\kern-.025em b}\kern-.08em
    T\kern-.1667em\lower.7ex\hbox{E}\kern-.125emX}}

\begin{document}

\title{Geo-RepNet: Geometry-Aware Representation Learning for Surgical Phase Recognition in Endoscopic Submucosal Dissection\\
}

\author{Rui Tang, Haochen Yin, Guankun Wang, Long Bai, An Wang, Huxin Gao, Jiazheng Wang, \\ Hongliang Ren,
\emph{Senior Member, IEEE}
\thanks{This work was supported by Hong Kong Research Grants Council (RGC) Collaborative Research Fund (C4026-21G), General Research Fund (GRF 14211420 \& 14203323), Shenzhen-Hong Kong-Macau Technology Research Programme (Type C) STIC Grant SGDX20210823103535014 (202108233000303). (\textit{Corresponding author: Hongliang Ren.})}
\thanks{R. Tang, H. Yin, G. Wang, L. Bai, A. Wang, H. Gao, and H. Ren are with the Department of Electronic Engineering, The Chinese University of Hong Kong, Hong Kong SAR, China.}
\thanks{J. Wang is with the Theory Lab, Central Research Institute, 2012 Labs, Huawei Technologies Co. Ltd., Hong Kong SAR, China}

}

\maketitle

\begin{abstract}
Surgical phase recognition plays a critical role in developing intelligent assistance systems for minimally invasive procedures such as Endoscopic Submucosal Dissection (ESD). However, the high visual similarity across different phases and the lack of structural cues in RGB images pose significant challenges. Depth information offers valuable geometric cues that can complement appearance features by providing insights into spatial relationships and anatomical structures. In this paper, we pioneer the use of depth information for surgical phase recognition and propose Geo-RepNet, a geometry-aware convolutional framework that integrates RGB image and depth information to enhance recognition performance in complex surgical scenes. Built upon a re-parameterizable RepVGG backbone, Geo-RepNet incorporates the Depth-Guided Geometric Prior Generation (DGPG) module that extracts geometry priors from raw depth maps, and the Geometry-Enhanced Multi-scale Attention (GEMA) to inject spatial guidance through geometry-aware cross-attention and efficient multi-scale aggregation. To evaluate the effectiveness of our approach, we construct a nine-phase ESD dataset with dense frame-level annotations from real-world ESD videos. Extensive experiments on the proposed dataset demonstrate that Geo-RepNet achieves state-of-the-art performance while maintaining robustness and high computational efficiency under complex and low-texture surgical environments.
\end{abstract}


\section{Introduction}
Endoscopic Submucosal Dissection (ESD) is a minimally invasive procedure for early-stage gastrointestinal cancer treatment that enables rapid en-bloc resection of large lesions while minimizing recurrence rates~\cite{wang2024copesd}. Automated surgical phase recognition is a fundamental component in developing intelligent surgical assistance systems for ESD, enabling surgical performance assessment, workflow monitoring, and real-time guidance to augment surgeon accuracy and mitigate procedural risks~\cite{bai2024ossar,czempiel2020tecno}. Despite significant advances, existing methodologies exhibit critical limitations: frame-level supervision introduces ambiguity when visually similar frames occur across distinct surgical phases~\cite{bai2025multimodal}, while computational constraints impede effective fusion of local and global temporal features~\cite{gao2021trans,ramesh2021multi}. Existing approaches mainly employ transformer-based temporal modeling~\cite{yi2021asformer}, which suffer from quadratic computational complexity and are prone to overfitting in surgical scenarios where visual similarity between phases is high and training data is limited~\cite{touvron2021training}. Additionally, the majority of current methods operate exclusively on RGB images~\cite{yuan2025recognizing}, neglecting valuable geometric information that could enhance surgical scene understanding.

Surgical phase recognition suffers from visual ambiguity and limited spatial context, where similar appearances across phases confound RGB-based methods. 
Depth information provides geometric priors that can effectively address these limitations by enabling robust discrimination of visually similar phases and better understanding of tool-tissue spatial relationships~\cite{ding2021repvgg,litjens2017survey}. 
Recent advances in depth estimation techniques have demonstrated significant potential for providing richer geometric information in surgical scene understanding~\cite{cui2024endodac,chen2020bi}. If effectively integrated as auxiliary information, depth cues could substantially enhance surgical phase recognition by addressing challenging conditions, including suboptimal illumination, specular reflections, and anatomical occlusion that commonly occur in endoscopic procedures.
However, existing RGB-D fusion approaches typically employ dual-encoder architectures that process depth information through dedicated networks~\cite{bodenstedt2019prediction,bai2025v}, introducing significant computational overhead and potential overfitting to incomplete geometric data. In surgical environments such as ESD, raw depth maps often suffer from missing texture details and inaccurate reconstruction due to occlusions and translucent tissues~\cite{cui2024surgical,liu2019dense}. Traditional RGB-D processing treats modalities equally, leading to computational redundancy when depth information is incomplete or noisy~\cite{fooladgar2020survey,marin2021token}, while failing to use depth as structural guidance.

Nevertheless, re-parameterizable architectures like RepVGG~\cite{ding2021repvgg} offer compelling solutions through multi-branch training that collapses into efficient single-path inference, showing particular promise in medical imaging with limited data~\cite{shen2017deep}. Recent multimodal fusion advances have explored attention-based mechanisms for adaptive feature weighting~\cite{ramachandran2019stand,wang2018non}, with geometric priors representing a shift toward direct attention guidance rather than explicit encoding~\cite{wang2022pvt,liu2021swin}.
However, existing methods have not explored integrating lightweight geometric priors with re-parameterizable architectures for surgical phase recognition. Current RGB-D approaches either suffer from computational inefficiency due to dual-stream processing or fail to effectively utilize depth as structural guidance rather than explicit features. This gap motivates the development of geometry-aware frameworks that are both efficient and tailored for surgical applications.

\begin{figure}[!t]
\centering
\includegraphics[width=0.4\textwidth]{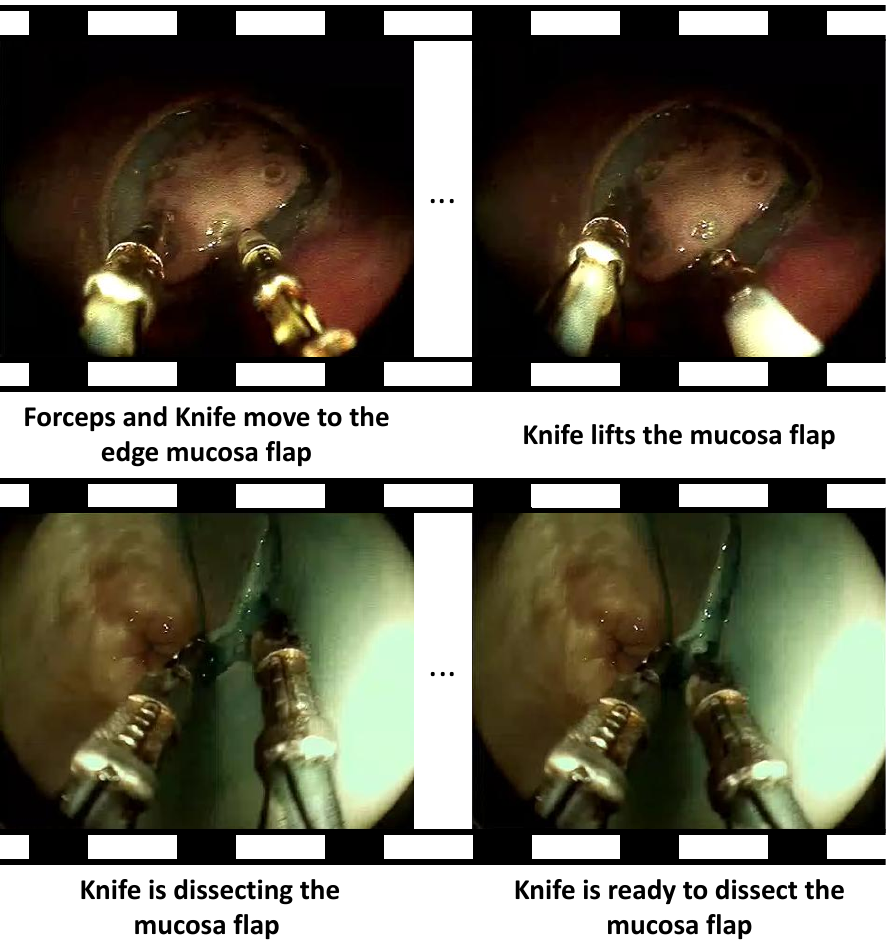}
\caption{Different surgical phases can exhibit high visual feature similarity, causing potential misclassification.}
\label{fig_tesla}
\end{figure}

To tackle the above challenges, we present \textbf{Geo-RepNet}, a geometry-aware convolutional framework tailored for surgical phase recognition in ESD. To address challenges posed by inter-phase visual similarity and limited spatial structure in RGB imagery, Geo-RepNet integrates a re-parameterizable RepVGG backbone with two key components. The \textbf{Depth-Guided Geometric Prior Generation (DGPG)} module encodes raw depth maps into geometric priors, which serve as spatial guidance without incurring the overhead of explicit depth encoding. The \textbf{Geometry-Enhanced Multi-scale Attention (GEMA)} further injects these priors into the feature extraction process via geometry-aware attention across multiple scales. This design enables efficient and effective RGB-D fusion while avoiding the computational cost of dual-stream architectures. Extensive experiments on real-world ESD data demonstrate that Geo-RepNet consistently surpasses both convolutional and transformer-based baselines, highlighting the effectiveness of incorporating geometric priors for efficient surgical scene understanding.



\section{Method}
\begin{figure*}[!t]
\centering
\includegraphics[width=0.85\textwidth]{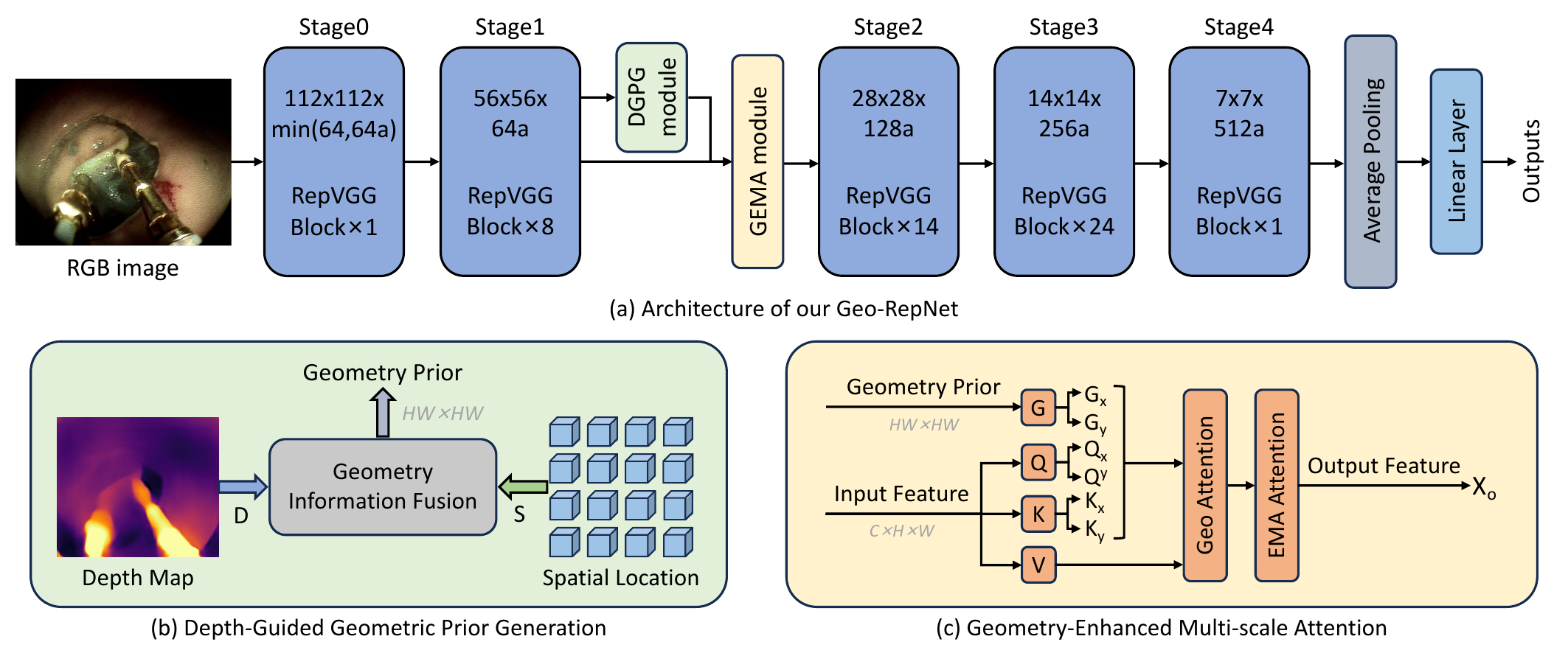}
\caption{Overview of our Geo-RepNet framework. (a) Architecture of Geo-RepNet; (b) Depth-Guided Geometric Prior Generation Module; (c) Geometry-Enhanced Multi-scale Attention Module.}
\label{fig_main}
\end{figure*}

In this work, we propose Geo-RepNet, a geometry-aware surgical phase recognition framework tailored for ESD procedures. As illustrated in Fig.~\ref{fig_main}, given an input surgical image $\mathbf{x} \in \mathbb{R}^{H \times W \times C}$ and its corresponding depth map $\mathbf{d} \in \mathbb{R}^{H \times W}$, our goal is to predict the surgical phase label $p \in \{1, 2, ..., K\}$. Unlike conventional models that rely solely on RGB appearance, Geo-RepNet incorporates depth as a geometric prior to providing structural cues such as spatial layout and object boundaries---features particularly useful in addressing the high inter-phase similarity and texture deficiency prevalent in endoscopic scenes.

To this end, Geo-RepNet adopts a convolutional backbone based on RepVGG, utilizing its re-parameterizable structure for efficient inference and strong generalization. We introduce the Depth-guided Geometric Prior Generation (DGPG) module, which derives relative positional encodings and visibility masks directly from raw depth, thereby enhancing spatial understanding without the overhead of explicit depth encoding. Furthermore, we propose the lightweight Geometry-Enhanced Multi-scale Attention (GEMA) block, which injects these geometric priors into attention mechanisms at multiple scales. Specifically, GEMA uses positional encodings to emphasize subtle spatial layout differences that distinguish visually similar surgical phases, while visibility masks direct attention away from texture-deficient regions—such as blood-filled or occluded areas—toward structurally informative regions like incision boundaries or tissue-tool interfaces. As a result, GEMA enhances discriminability under high inter-phase ambiguity and compensates for missing texture details by focusing on geometric and structural cues.

\subsection{Overview of Geo-RepNet architecture}

As illustrated in Fig.~\ref{fig_main}(a), we build Geo-RepNet on the convolutional backbone, RepVGG, to achieve efficient and robust surgical phase recognition. Compared to Transformer-based architectures, which are computationally intensive and prone to overfitting in surgical scenes with high inter-phase similarity and low-texture regions, convolutional models offer a stronger inductive bias for modeling local spatial patterns. RepVGG further enhances these benefits with its re-parameterizable design, enabling fast inference and strong generalization under limited training data. Its compact structure also facilitates real-time deployment in resource-constrained surgical settings.

In surgical environments, raw depth maps often contain incomplete geometry due to occlusions, specular reflections, and translucent tissues, leading to missing texture details and reconstruction errors. Directly feeding such noisy depth into convolutional backbones can introduce unreliable signals and increase the computational load. To mitigate this, we transform depth into lightweight geometric priors that guide spatial attention, rather than encoding depth features directly. This approach retains structural cues while avoiding overfitting to noisy depth and reducing memory usage.

Geo-RepNet builds upon a RepVGG backbone for hierarchical feature extraction, enhanced by two auxiliary modules: the Depth-Guided Geometric Prior Generation (DGPG) module, which derives spatial priors from depth, and the Geometry-Enhanced Multi-Scale Attention (GEMA) block, which integrates these priors into the feature learning process.

Given an input RGB image $\mathbf{x} \in \mathbb{R}^{H \times W \times 3}$ and its corresponding depth map $\mathbf{d} \in \mathbb{R}^{H \times W}$, we insert the DGPG and GEMA modules after Stage 1 of the RepVGG backbone. This location offers a balance between spatial resolution and semantic abstraction: earlier stages lack sufficient context, while deeper stages suffer from excessive downsampling. Injecting geometric priors at this stage preserves local structure (e.g., tool-tissue boundaries) and facilitates the disambiguation of visually similar regions.

From the raw depth map, DGPG computes relative positional encodings and visibility masks:
\begin{equation}
(\mathbf{r}_{\sin}, \mathbf{r}_{\cos}), \mathbf{m}_{\mathrm{vis}} = \mathcal{P}(\mathbf{d}; H_1, W_1)
\end{equation}
where $(H_1, W_1)$ denotes the resolution of the intermediate feature map $\mathbf{f}_1$, and $\mathcal{P}(\cdot)$ is defined in Section~\ref{sec:DGPG}.

These priors are injected into the GEMA block, which applies geometry-aware, multi-scale attention to produce enhanced features:
\begin{equation}
\hat{\mathbf{f}}_1 = \mathrm{GEMA}(\mathbf{f}_1, \mathbf{r}_{\sin}, \mathbf{r}_{\cos}, \mathbf{m}_{\mathrm{vis}})
\end{equation}

The refined representation $\hat{\mathbf{f}}_1$ is then forwarded through the remaining stages of the RepVGG backbone. This early-stage integration improves the network’s ability to model structural dependencies and distinguish subtle phase transitions while maintaining enhanced interpretability.

\subsection{Depth-Guided Geometric Prior Generation}
\label{sec:DGPG}

As illustrated in Fig.~\ref{fig_main}(b), to address the challenges of low-texture regions, high visual similarity across surgical phases, and limited annotated data, we propose the Depth-Guided Geometric Prior Generation (DGPG) module to enable global structural reasoning beyond local receptive fields. Instead of relying solely on deep convolutional layers, DGPG extracts long-range spatial dependencies by encoding relative position and visibility from low-dimensional depth and coordinate grids. Inspired by recent advances in lightweight geometric encoding~\cite{yin2025dformerv2}, this design avoids the computational burden of conventional depth-processing backbones and is well-suited for real-time deployment in surgical settings.

Given an input depth map $\mathbf{d} \in \mathbb{R}^{B \times 1 \times H \times W}$, the module first interpolates it to the target resolution and computes \textit{relative depth differences} between spatial positions. For each pair of pixels $(i,j)$ and $(i',j')$, the absolute depth difference $|\mathbf{d}_{i,j} - \mathbf{d}_{i',j'}|$ is scaled by a learnable, head-specific exponential decay factor defined as:
\begin{equation}
\text{decay}(h) = \log\left(1 - 2^{-(\lambda_0 + \gamma \cdot \frac{h}{H})} \right)
\end{equation}
where $h$ indexes the attention head, and $\lambda_0$, $\gamma$ are learnable parameters controlling the initial decay and scaling behavior across heads. In parallel, we generate sinusoidal \textit{relative positional encodings} using a fixed set of frequencies:
\begin{equation}
\text{PE}_{\sin}(i,j) = \sin(\Delta_{ij} \cdot \omega), \quad
\text{PE}_{\cos}(i,j) = \cos(\Delta_{ij} \cdot \omega)
\end{equation}
where $\Delta_{ij}$ denotes the relative index between locations, and $\omega$ represents the frequency basis distributed over embedding dimensions. The final geometric prior comprises two components: an angular positional basis $(\text{PE}_{\sin}, \text{PE}_{\cos})$, which encodes relative spatial positions using sinusoidal embeddings in angular form, and a decay mask $\mathbf{M}_{\text{geo}}$ that provides spatial attention weights by combining position-based and depth-based distances:
\begin{equation}
\mathbf{M}_{\text{geo}} = w_1 \cdot \mathbf{M}_{\text{pos}} + w_2 \cdot \mathbf{M}_{\text{depth}}
\end{equation}
where $w_1$ and $w_2$ are learnable fusion weights, and the two terms represent decay masks derived from coordinate and depth differences, respectively.

The DGPG module supports both 1D (along the height and width axes separately) and 2D (full spatial map) formulations, enabling flexible encoding of structural priors. The generated geometric prior \(G = \{G_x, G_y\}\) is shared across attention heads to modulate the pairwise interaction strength based on geometric affinity. By capturing both relative positional offsets and local depth variations, DGPG introduces a lightweight yet expressive structural prior that enhances the model’s capacity for spatial reasoning in complex and deformable surgical environments. Accurate modeling of such geometric structure is crucial for distinguishing visually similar surgical phases, compensating for texture-deficient regions, and improving robustness under limited training data.

\subsection{GEMA: Geometry-Enhanced Multi-scale Attention}

As illustrated in Figure~\ref{fig_main}(c), the proposed Geometry-Enhanced Multi-scale Attention module consists of two key components: geometry-aware cross self-attention (CrossGSA)~\cite{yin2025dformerv2} and efficient multi-scale attention (EMA)~\cite{Ouyang_2023}.

Given input features \( X \in \mathbb{R}^{B \times C \times H \times W} \) and geometric prior \( G \in \mathbb{R}^{H \times W \times 2} \), the Geometry-Enhanced Multi-scale Attention (GEMA) module first computes attention embeddings as:
\begin{align}
Q &= W_q X, \quad K = W_k X, \\
V &= \text{DWConv}(W_v X), \quad G_x, G_y = \mathcal{T}(G)
\end{align}

where \( W_q, W_k, W_v \) are learnable linear projections, \(\text{DWConv}\) denotes depthwise convolution for local enhancement, and \(\mathcal{T}(\cdot)\) applies sinusoidal positional encoding to the geometry prior. The geometry-aware transformation is then applied to inject positional priors into the attention computation:
\begin{align}
Q_x, Q_y &= \mathcal{T}(Q, \sin, \cos), \quad K_x, K_y = \mathcal{T}(K, \sin, \cos), \\
A &= \text{Softmax} \left( \frac{Q_x K_x^\top + Q_y K_y^\top}{\sqrt{d_k}} + M \right), \\
O &= X + W_o (A V) + \text{LayerNorm}(\cdot)
\end{align}
where \( M \) is an optional geometry bias map, and \( W_o \) is an output projection matrix. The EMA further captures spatial and channel-wise context by grouping input channels and applying directional pooling:
\begin{equation}
X_h = \text{AvgPool}_h(X), \quad X_w = \text{AvgPool}_w(X)
\end{equation}
The pooled features \([X_h; X_w]\) are then refined by a series of \(1 \times 1\) and \(3 \times 3\) convolutions, followed by group normalization. The resulting map is activated using a sigmoid function to generate attention weights:
\begin{equation}
W = \sigma(Z), \quad \tilde{X} = X \odot W
\end{equation}
where \(\sigma(\cdot)\) denotes the sigmoid function and \(\odot\) indicates element-wise multiplication. Overall, GEMA effectively fuses RGB features with geometry priors through geometry-aware attention and multi-scale spatial modeling, enabling robust and efficient representation learning in complex surgical scenes.

\section{Experiment}
\subsection{Datasets}
We construct a phase classification dataset tailored to the ESD procedure, encompassing nine clinically relevant phases annotated from real-world surgical videos. Each frame is assigned one of the following categories: (1) \textit{Forceps grasp}, (2) \textit{Forceps lift the mucosa flap}, (3) \textit{Forceps and Knife move to the edge mucosa flap}, (4) \textit{Forceps rotate}, (5) \textit{Knife lifts the mucosa flap}, (6) \textit{Adjust the camera position}, (7) \textit{Clean the camera lens}, (8) \textit{Knife is dissecting the mucosa flap}, and (9) \textit{Knife is ready to dissect the mucosa flap}. The dataset is split into a training set and a validation set, comprising 7080 and 792 samples, respectively. The label distribution exhibits class imbalance, with prediction-relevant stages (e.g., classes 8 and 9) occupying a significantly larger proportion. Table~\ref{tab:surgical_phase_samples} summarizes the sample distribution across the two splits. This realistic imbalance poses additional challenges for reliable phase recognition in complex surgical scenes.


\subsection{Implementation Details}
We evaluate our proposed classification framework on the ESD phase recognition dataset. All models are trained for 150 epochs using the Adam optimizer with a batch size of 16. The learning rate is initialized to $1 \times 10^{-4}$ and scheduled via cosine annealing. Training and evaluation are conducted using the PyTorch framework on an NVIDIA RTX 4090 GPU.

We compare our method against a diverse set of baseline architectures, categorized into convolutional and transformer-based models.
For convolutional architectures, we evaluate MobileViTV3~\cite{wadekar2022mobilevitv3mobilefriendlyvisiontransformer}, ConvNeXt~\cite{liu2022convnet2020s}, FastNet~\cite{chen2023rundontwalkchasing}, and Res2Net-50~\cite{Gao_2021}. 
For transformer-based architectures, we consider VMamba~\cite{liu2024vmambavisualstatespace}, SwinTransformerV2~\cite{liu2022swintransformerv2scaling}, Twins~\cite{chu2021twinsrevisitingdesignspatial}, UniFormer~\cite{li2023uniformerunifyingconvolutionselfattention}, and CSWinTransformer~\cite{dong2022cswintransformergeneralvision}.
Additionally, we include MambaOut~\cite{yu2024mambaoutreallyneedmamba}. Performance is measured using three standard classification metrics: Accuracy (AC), F-score (F1), and Area Under the Curve (AUC), which collectively capture model effectiveness in handling class imbalance and fine-grained visual distinctions across phases.
\begin{table}
\centering
\caption{Sample Distribution of Surgical Phase Classification Dataset}
\begin{tabular}{l c c}
\toprule
\textbf{ESD Phase Name} & \textbf{Train} & \textbf{Test} \\
\midrule
Forceps grasp & 81 & 9 \\
Forceps lift the mucosa flap & 504 & 57 \\
Forceps and Knife move to the edge mucosa flap & 444 & 51 \\
Forceps rotate & 21 & 3 \\
Knife lifts the mucosa flap & 405 & 48 \\
Adjust the camera position & 333 & 39 \\
Clean the camera lens & 303 & 36 \\
Knife is dissecting the mucosa flap & 2628 & 294 \\
Knife is ready to dissect the mucosa flap & 2361 & 264 \\
\bottomrule
\end{tabular}
\label{tab:surgical_phase_samples}
\end{table}
\subsection{Results}
\subsubsection{Evaluation on the ESD Dataset} 
Table~\ref{tab:backbone_comparison_sorted} presents the performance of different backbones on the surgical phase recognition task. Among all CNN-based models, our method achieves the highest accuracy of 85.02\%, F1-score of 81.74\%, and AUC of 93.10\%, demonstrating superior capability in modeling complex surgical scenes. Res2Net50 achieves an accuracy of 75.66\%, while ConvNeXt reaches 74.53\%, both significantly lower than our model across all evaluation metrics. Lightweight models such as FastNet and MobileViT-v3 achieve accuracies of 75.54\% and 73.02\%, respectively, but still lag behind in terms of F1-score and AUC. Vision Transformer-based methods like SwinTransformerV2 and UniFormer show moderate performance, yet their F1-score and AUC remain inferior to our approach. Notably, MambaOut and V-Mamba yield lower accuracy and F1-score, indicating that generic sequence modeling backbones may not effectively capture the spatial and temporal dependencies specific to surgical procedures. These results indicate that our backbone design is better suited for surgical scene understanding, providing consistent improvements in capturing task-specific semantics and structural priors.

\begin{table}[]
\centering
\caption{Comparison of classification performance across different backbones (sorted by Accuracy). All metrics are reported in percentages.}
\begin{tabular}{lccc}
\toprule
\textbf{Backbone} & \textbf{Accuracy (\%)} & \textbf{F1 (\%)} & \textbf{AUC (\%)} \\
\midrule
Twins            & 63.69 & 65.30 & 75.41 \\
MambaOut         & 68.16 & 62.35 & 77.55 \\
Vmanba            & 69.03 & 64.09 & 80.67 \\
SwinTransformerV2 & 70.41 & 66.67 & 81.41 \\
MobileViTV3       & 73.02 & 74.97 & 80.44 \\
UniFormer         & 73.90 & 77.89 & 77.19 \\
ConvNeXt          & 74.53 & 76.19 & 83.67 \\
FastNet          & 75.54 & 74.88 & 77.79 \\
Res2Net-50         & 75.66 & 76.29 & 82.53 \\
CSWinTransformer     & 77.01 & 78.56 & 84.53 \\
Ours              & \textbf{85.02} & \textbf{81.74} & \textbf{93.10} \\
\bottomrule
\end{tabular}
\label{tab:backbone_comparison_sorted}
\end{table}

\subsubsection{Ablation Studies}
Table~\ref{tab:ablation studies} reports the results of an ablation study conducted to assess the individual and combined contributions of three key components in our framework: DGPG (Depth-Guided Geometric Prior Generation), GSA (geometry-aware cross self-attention), and EMMA (Efficient Multi-scale Modality-aware Attention). The baseline model without any of these modules achieves an accuracy of 77.53\%, an F1-score of 71.88\%, and an AUC of 83.74\%. Introducing EMMA alone improves the overall performance, especially in terms of AUC (88.07\%), demonstrating its effectiveness in capturing multi-scale contextual information. The addition of DGPG or GSA  individually also leads to performance gains, suggesting that depth and spatial priors both contribute meaningful structural cues for classification. Combining DGPG with GSA yields further improvements across all metrics, indicating complementary effects of depth-aware and spatial-aware representations. When EMMA is combined with either DGPG or GSA, we observe substantial gains in F1-score and AUC, validating the synergy between attention mechanisms and structural priors. Finally, the full model incorporating all three components, DGPG, GSA, and EMMA, achieves the best performance, with an F1-score of 81.74\%, accuracy of 85.02\%, and AUC of 93.10\%. These results confirm that the joint modeling of depth, spatial priors, and multi-scale attentional features significantly improves the performance in complex surgical scenes.

\begin{table}[]
\centering
\caption{Ablation Study on Classification Performance}
\begin{tabular}{lccc}
\toprule
\textbf{Method} & \textbf{Acc. (\%)} & \textbf{F1 (\%)} & \textbf{AUC (\%)} \\
\midrule
None                 & 77.53 & 71.88 & 83.74 \\
EMMA                 & 78.65 & 74.66 & 88.07 \\
DGPG                 & 80.75 & 73.45 & 82.83 \\
GSA                   & 80.15 & 79.05 & 85.56 \\
DGPG + GSA            & 81.65 & 75.49 & 90.47 \\
DGPG + EMMA          & 83.90 & 81.05 & 86.24 \\
GSA  + EMMA           & 82.40 & 80.07 & 92.08 \\
DGPG + GSA  + EMMA    & \textbf{85.02} & \textbf{81.74} & \textbf{93.10} \\
\bottomrule
\end{tabular}
\label{tab:ablation studies}
\end{table}

\subsubsection{Fine-tuning Experiments}
Table~\ref{tab:factor_ablation} reports the phase recognition results of our Geo-RepNet model under varying grouping factors in the EMA module. The best performance is observed when the factor is set to 4, with an accuracy of 85.02\%, F1-score of 81.74\%, and AUC of 93.10\%. As the grouping factor increases from 4 to 32, the model exhibits a slight performance decline. Nonetheless, even with a larger factor of 32, Geo-RepNet still achieves competitive results, reaching an accuracy of 83.90\%, F1-score of 79.84\%, and AUC of 90.82\%, outperforming all baseline methods reported in Table~\ref{tab:backbone_comparison_sorted}. This demonstrates that our framework maintains strong robustness against internal architectural variations. The integration of geometric prior guidance further enhances its ability to deliver consistent and generalized performance, which is particularly critical in surgical scenarios characterized by subtle and highly variable visual cues.

\begin{table}[t]
\centering
\caption{Effect of Different Grouping Factors on Classification Performance}
\begin{tabular}{lccc}
\toprule
\textbf{Factor} & \textbf{Accuracy (\%)} & \textbf{F1 (\%)} & \textbf{AUC (\%)} \\
\midrule
4  & 85.02 & 81.74 & 93.10 \\
8  & 84.40 & 80.98 & 90.42 \\
16 & 84.77 & 78.78 & 91.57 \\
32 & 83.90 & 79.84 & 90.82 \\
\bottomrule
\end{tabular}
\label{tab:factor_ablation}
\end{table}

\section{Conclusion}
In this work, we introduced Geo-RepNet, a geometry-aware representation
learning framework for surgical phase recognition in endoscopic submucosal dissection. By incorporating depth-guided geometric priors through the proposed DGPG and GEMA modules, our approach effectively addresses visual ambiguities and structural deficiencies inherent in endoscopic scenes. Built on a re-parameterizable RepVGG backbone, Geo-RepNet achieves superior accuracy and efficiency compared to both convolutional and transformer-based baselines. Extensive experiments and ablation studies confirm the effectiveness of using lightweight geometric guidance for robust, real-time phase recognition. This work highlights the potential of integrating depth-aware structural reasoning into surgical video understanding and establishes a foundation for developing more effective surgical assistance systems.

\bibliographystyle{IEEEtran}
\bibliography{reference}

\end{document}